\documentclass[sigconf,nonacm]{acmart}

\newcommand{\etal}{et~al.\xspace}
\newcommand{\ie}{i.e.\xspace}
\newcommand{\eg}{e.g.\xspace}

\newcommand{\conclusionbox}[2]{%
    \begin{mdframed}[%
        skipabove=3pt,
        skipbelow=3pt,
        innertopmargin=3pt,
        innerleftmargin=3pt,
        innerrightmargin=3pt,
        innerbottommargin=3pt
      ]
        \noindent {\bf Conclusion for #1.} #2
    \end{mdframed}
}

\acmConference[E-QSE 2025]{%
    The 2nd International Workshop on Empirical Studies for Quantum Software
    Engineering
}{%
    17--20 June, 2025
}{%
    Istanbul, Türkiye
}

\title[QAOA-PCA: Enhancing Efficiency in QAOA via PCA]{%
    QAOA-PCA: Enhancing Efficiency in the Quantum Approximate Optimization
    Algorithm via Principal Component Analysis
}

\author{Owain Parry}
\affiliation{\institution{University of Sheffield}\country{UK}}
\author{Phil McMinn}
\affiliation{\institution{University of Sheffield}\country{UK}}

\begin{abstract}
    The Quantum Approximate Optimization Algorithm (QAOA) is a promising
    variational algorithm for solving combinatorial optimization problems on
    near-term devices.
    However, as the number of layers in a QAOA circuit increases, which is
    correlated with the quality of the solution, the number of parameters to
    optimize grows linearly.
    This results in more iterations required by the classical optimizer, which 
    results in an increasing computational burden as more circuit executions are 
    needed.
    To mitigate this issue, we introduce QAOA-PCA, a novel reparameterization 
    technique that employs Principal Component Analysis (PCA) to reduce the 
    dimensionality of the QAOA parameter space.
    By extracting principal components from optimized parameters of smaller 
    problem instances, QAOA-PCA facilitates efficient optimization with fewer 
    parameters on larger instances.
    Our empirical evaluation on the prominent MaxCut problem demonstrates that 
    QAOA-PCA consistently requires fewer iterations than standard QAOA, 
    achieving substantial efficiency gains.
    While this comes at the cost of a slight reduction in approximation ratio 
    compared to QAOA with the same number of layers, QAOA-PCA almost always 
    outperforms standard QAOA when matched by parameter count.
    QAOA-PCA strikes a favorable balance between efficiency and performance,
    reducing optimization overhead without significantly compromising
    solution quality.
\end{abstract}



\thanks{%
    The authors are supported by the EPSRC grant "Test FLARE" (EP/X024539/1) and 
    the Robust and Reliable Quantum Computing Programme (RoaRQ)
}

\usepackage{xspace}
\usepackage{booktabs}
\usepackage{colortbl}
\usepackage{subcaption}
\usepackage{xurl}
\usepackage{bm}
\usepackage{mdframed}

\begin{document}
    \maketitle
    \section{Introduction}
\label{sec:introduction}

Quantum computing leverages quantum mechanical phenomena to perform computations 
that are intractable on classical systems~\cite{Nielsen2010,Preskill2018}.
One prominent application is combinatorial optimization, where the Quantum 
Approximate Optimization Algorithm (QAOA) is a promising approach to tackle 
problems like MaxCut~\cite{Farhi2014}.
The goal in MaxCut is to partition a graph's vertices to maximize the number of
edges crossing the partition in the unweighted case, or to maximize the total
weight of these edges in the weighted case.
It finds important applications in diverse fields~\cite{Wang2010}.

QAOA operates by applying \(p\) layers of alternating cost and mixing
Hamiltonians parameterized by two angles, with each layer increasing solution 
quality~\cite{Galda2021}.
A major challenge in the practical deployment of QAOA is the optimization of its
\(2p\) parameters.
This involves a classical-quantum loop where a classical optimizer proposes
parameter values used to execute the QAOA circuit on a quantum computer to
evaluate the cost function at each iteration.
This becomes more expensive as the number of layers, and thus parameters, 
increases.
The burden of repeated circuit executions, and the non-convex cost 
landscape~\cite{Lyngfelt2025}, contributes to the quantum software engineering 
challenge of scaling QAOA to larger problem instances (\eg, graphs with more 
vertices), especially for weighted MaxCut due to the proliferation of local 
minima~\cite{Sureshbabu2024}.

Recent research has identified parameter concentration, where optimal parameters
cluster around specific values, and transferability, where parameters 
optimized for one instance are applicable to similar instances.
This suggests redundancy in the parameter space, implying that reduced
dimensionality could improve efficiency without sacrificing
performance~\cite{Akshay2021,Zeng2024,Brandao2018,Galda2023,Galda2021,
Shaydulin2023,Montanez2024,Shi2022}.
Building on this, we propose QAOA-PCA, a novel reparameterization technique to 
improve the efficiency of QAOA.
Rather than optimizing the full QAOA parameter set, QAOA-PCA leverages 
Principal Component Analysis (PCA)~\cite{Abdi2010} to extract key features from 
the optimal parameters of a training set of smaller instances.
These principal components form a lower-dimensional subspace, resulting in fewer 
parameters to optimize when tackling larger instances, leading to faster 
convergence and fewer circuit executions.

We conducted an extensive empirical evaluation of QAOA-PCA supported by robust 
statistical testing.
We collected all connected, non-isomorphic graphs on 5--7 vertices to form an 
unweighted training set of 986 graphs.
To create a weighted training set, we assigned random edge weights to these.
We ran QAOA with 2, 4, and 8 layers on all graphs in both training sets.
For these six configurations of circuit depth and training set, we applied PCA 
to extract the principal components from the optimized parameters.
We sampled 1,000 8-vertex weighted graphs to form an evaluation set.
Instead of optimizing the full QAOA parameters for these graphs, we optimized
the coefficients of the most important principal components as substitute
parameters.
We did this for various numbers of components for each of the six training 
configurations.
We ensured at least a 50\% reduction in the number of parameters compared to
standard QAOA with the same number of layers, which we used as a baseline.
As another baseline, we used standard QAOA with the same number of parameters as
components and thus fewer layers.

Our results demonstrate that QAOA-PCA requires significantly fewer iterations, 
and thus circuit executions, compared to standard QAOA of the same depth.
With the same number of parameters, QAOA-PCA generally requires slightly fewer.
QAOA-PCA achieves only a slightly reduced approximation 
ratio~\cite{Schwagerl2024} compared to standard QAOA with the
same number of layers.
It almost always outperforms standard QAOA when matched by parameter count.
By broadly preserving the advantage of deeper QAOA circuits while avoiding 
excessive optimization overhead, QAOA-PCA achieves a favorable trade-off between
efficiency and performance.

The main contributions of this study are as follows: \\
{\bf Contribution 1: QAOA-PCA.} We introduce QAOA-PCA, a novel
reparameterization technique that leverages PCA to reduce the dimensionality of
the QAOA parameter space, improving efficiency. \\
{\bf Contribution 2: Empirical Evaluation.} We provide a rigorous empirical
evaluation of QAOA-PCA compared to standard QAOA. \\
{\bf Contribution 3: Results.} We demonstrate that QAOA-PCA achieves a favorable
trade-off between performance and efficiency.

    \section{Background}
\label{sec:background}

{\bf The Quantum Approximate Optimization Algorithm (QAOA)} is a 
quantum-classical algorithm for solving combinatorial optimization 
problems~\cite{Farhi2014}.
It employs a parameterized circuit of depth \(p\) alternating between unitaries 
generated by the cost Hamiltonian \(H_C\) (encoding the problem) and the mixing 
Hamiltonian \(H_B\) (facilitating computational state transitions).
The resulting state is
\[
|\psi(\bm{\gamma}, \bm{\beta})\rangle = \prod_{i=1}^p e^{-i\beta_i H_B} \, e^{-i\gamma_i H_C} \, |+\rangle^{\otimes n},
\]
where \(|+\rangle^{\otimes n}\) denotes a uniform superposition over all
computational basis states, and \(\bm{\gamma} = (\gamma_1,\dots,\gamma_p)\) and
\(\bm{\beta} = (\beta_1,\dots,\beta_p)\) are the variational parameters.
These parameters are optimized classically to minimize the expectation value of
the cost Hamiltonian:
\[
(\bm{\gamma}^*, \bm{\beta}^*) = \arg\min_{\bm{\gamma}, \bm{\beta}} \langle \psi(\bm{\gamma}, \bm{\beta}) | H_C | \psi(\bm{\gamma}, \bm{\beta})\rangle.
\]

{\bf MaxCut} aims to cut (partition) the vertices of a graph \(G=(V,E)\) into 
two sets to maximize the cut value, \ie, the total weight of the edges between 
them.
A cut is represented by \(\bm{z} = (z_1,\ldots,z_n)\), with
\(z_i \in \{+1,-1\}\) indicating the set of vertex \(i\).
The cut value of \(\mathbf{z}\) is
\[
C(\mathbf{z}) = \sum_{(i,j)\in E} w_{ij}\frac{1-z_i z_j}{2},
\]
where \(w_{ij}\) is the weight of edge \((i,j)\), or 1 in the unweighted case.
The ground state of \(H_C\) corresponds to the optimal cut.
QAOA performance is measured by the approximation ratio
\[
r(\bm{\gamma}, \bm{\beta}) = \frac{\langle \psi(\bm{\gamma}, \bm{\beta}) | H_C | \psi(\bm{\gamma}, \bm{\beta}) \rangle}{C_{\text{min}}},
\]
where \(C_{\text{min}}\) is the minimum energy of \(H_C\), corresponding to the
maximum cut value.
This ratio quantifies how close the obtained solution is to the optimal one: a
value near 1 indicates near-optimal performance, while a lower value reflects a
suboptimal cut.

{\bf Trotterized Quantum Annealing (TQA)} provides a heuristic for initializing 
QAOA parameters.
In quantum annealing, the system is gradually driven from \(H_B\) to \(H_C\) 
over a total time \(T\).
Discretizing this evolution into \(p\) steps of duration \(\Delta t = T/p\) via
the Suzuki-Trotter decomposition produces the initial QAOA parameters
\[
\gamma_i = \frac{i}{p}\Delta t, \quad \beta_i = \left(1-\frac{i}{p}\right)\Delta t,\quad i=1,\dots,p.
\]
An appropriate choice of \(\Delta t\) can guide the optimization towards the
global minimum while avoiding poor local minima~\cite{Sack2021}.

{\bf Principal Component Analysis (PCA)} reduces the dimensionality of 
a dataset while retaining most of its variance~\cite{Abdi2010}.
Given a \(n \times m\) mean-centered data matrix \(X\), the covariance matrix is
\(\Sigma = \frac{1}{n-1} X^\top X\).
The principal components are the eigenvectors associated with the largest 
eigenvalues of \(\Sigma\).

    \section{Approach}
\label{sec:approach}

QAOA shows promise in addressing combinatorial optimization 
problems~\cite{Farhi2014}.
Its application is hindered by the need to optimize numerous parameters as the 
number of layers increases.
We introduce QAOA-PCA to mitigate this, a reparameterization technique 
leveraging PCA to exploit redundancies in QAOA parameters.

Recent research has identified parameter concentration and transferability.
Concentration is the tendency of optimal QAOA parameters to cluster around 
specific values as the problem size (\eg, vertex count) increases, indicating 
that parameters optimized on smaller problem instances can be effective on 
larger ones~\cite{Akshay2021,Galda2023,Zeng2024,Brandao2018}.
Transferability denotes the applicability of parameters optimized for one
instance to different instances, which is predictable from graph properties when 
solving MaxCut~\cite{Galda2023,Galda2021,Shaydulin2023,Montanez2024}.
These insights imply that the effective dimensionality of the parameter space is
lower than its nominal size~\cite{Shi2022}, presenting an opportunity to reduce
the number of parameters requiring optimization.

QAOA-PCA capitalizes on concentration and transferability by employing PCA to
identify the principal components of optimal QAOA parameters from a training set
of smaller problem instances.
By efficiently narrowing the search space, QAOA-PCA reduces computational cost
while maintaining high solution quality when addressing larger instances.
The process involves four stages: \\
{\bf Stage 1: Data Collection.} Gather optimized QAOA parameters from a diverse
set of smaller problem instances. \\
{\bf Stage 2: Principal Component Analysis.} Apply PCA to the optimized
parameters to identify the principal components that capture the most
significant variations in the parameter space. \\
{\bf Stage 3: Reparameterization.} Represent the QAOA parameters of new, larger
problem instances as linear combinations of the identified principal components.
This approach reconstructs the full parameter vector from a much smaller set of
variables. \\
{\bf Stage 4: Optimization.} Optimize only the coefficients of these principal
components, thereby reducing the dimensionality of the optimization problem and
facilitating convergence.

    \section{Evaluation}
\label{sec:evaluation}

This section details the methodology of our empirical evaluation aimed at
addressing our research questions regarding QAOA-PCA: \\
{\bf RQ1: Efficiency.} How does the number of optimizer iterations required by
QAOA-PCA compare with that of standard QAOA? \\
{\bf RQ2: Performance.} How does the approximation ratio achieved by QAOA-PCA
compare with that of standard QAOA?

We created a Python script to automate all aspects of our empirical evaluation.
We make this available in our replication package~\cite{ReplicationPackage}.

\vspace{1mm} \noindent {\bf Training.}
We collected all connected, undirected, non-isomorphic graphs on 5, 6, and 7
vertices to form an unweighted training set of 986 graphs~\cite{McKay1983}.
Our script applied random edge weights to every graph to form a weighted 
training set.
It ran QAOA with 2, 4, and 8 layers on all graphs in both training sets
using ideal quantum simulation provided by Qiskit~\cite{Javadi2024}, 
initializing the parameters using TQA, and optimizing them using 
COBYLA~\cite{Pellow2021}.
For each combination of circuit depth and graph, our script repeated the process 
for five values of TQA time step \(\Delta t\) (0.1, 0.3, 0.5, 0.7, 0.9) and 
retained only the results of the run with the greatest approximation ratio.
For the six configurations of training set and circuit depth \(p\), our script 
applied PCA to extract the principal components from the \(986 \times 2p\) 
matrix of the optimized parameters of every graph.

\vspace{1mm} \noindent {\bf Evaluation.}
Our script randomly sampled 1,000 connected, undirected, non-isomorphic graphs 
on 8 vertices and applied random edge weights to form an evaluation set.
It ran QAOA on these as before with 1, 2, 4, and 8 layers.
This represents standard QAOA.
For each of the six sets of principal components that it extracted earlier, our 
script ran QAOA on the evaluation graphs again.
This time, it optimized the coefficients of the most important principal 
components as substitute parameters, trying five random initializations.
This represents QAOA-PCA with 2, 4, and 8 layers as trained on the optimal
parameters of both unweighted and weighted graphs.
In each case, our script repeated the process for different numbers of principal 
components, ensuring at least a 50\% reduction in the number of parameters 
compared to standard QAOA with the same number of layers.
For QAOA-PCA with 2 layers, it tried 2 principal components.
For 4 layers, it tried 2 and 4 components.
For 8 layers, it tried 2, 4, and 8.
This resulted in 12 configurations of QAOA-PCA.

\vspace{1mm} \noindent {\bf Comparison.}
We compared each configuration of QAOA-PCA against two baselines.
The first was against standard QAOA with the same number of layers and thus
twice as many parameters.
The second was against standard QAOA with the same number of parameters as
principal components and thus half as many layers.
For each comparison, our script performed two-tailed Wilcoxon signed-rank tests
with rank-biserial correlation (RBC) effect sizes.
It did this regarding the number of optimizer iterations and the approximation 
ratio to compare both the efficiency and performance respectively of QAOA-PCA to 
standard QAOA.
To calculate approximation ratio, our script obtained \(C_{\text{min}}\) by 
brute-force.

\vspace{1mm} \noindent {\bf Threats.}
Our results may not reflect the behavior of QAOA-PCA on real quantum computers
due to our use of ideal simulation.
However, we also used it for the standard QAOA baseline, so the comparison is
fair.
It is possible that our script did not converge on the globally optimal set of 
parameters when performing QAOA.
We mitigated this risk by trying multiple initializations, which increases the
probability.
Our results could be biased by our choice of optimizer, COBYLA.
We mitigated this by selecting an optimizer that is popular and well-regarded in
this domain~\cite{He2024,Campbell2022,Hao2024}.

    \section{Results}
\label{sec:results}

\begin{table*}[t]
    \setlength{\tabcolsep}{2.9pt}
    \centering
    \caption{%
        \label{tab:results}
        Comparison of QAOA-PCA and standard QAOA across 12 configurations.
        For each, the table gives the median (Med.) number of optimizer 
        iterations (RQ1) and approximation ratio (RQ2) for QAOA-PCA and the two 
        standard QAOA baselines (Same \# Layers, Same \# Param.).
        Wilcoxon test results include p-values (P-Val.) and rank-biserial 
        correlation (RBC) effect sizes.
    }
    \begin{tabular}{lrrrrrrrrrrrrrrrr}
        \toprule
        & & & \multicolumn{7}{c}{\bf Number of Iterations (RQ1)} &
        \multicolumn{7}{c}{\bf Approximation Ratio (RQ2)} \\
        \cmidrule(r){4-10}
        \cmidrule(l){11-17}
        & & & & \multicolumn{3}{c}{\bf Same \# Layers} & 
        \multicolumn{3}{c}{\bf Same \# Param.} &
        & \multicolumn{3}{c}{\bf Same \# Layers} &
        \multicolumn{3}{c}{\bf Same \# Param.} \\
        \cmidrule(r){5-7}
        \cmidrule(l){8-10}
        \cmidrule(r){12-14}
        \cmidrule(l){15-17}
        {\bf Training Set} &
        {\bf \# Layers} &
        {\bf \# Param.} &
        {\bf Med.} &
        {\bf Med.} &
        {\bf P-Val.} &
        {\bf RBC} &
        {\bf Med.} &
        {\bf P-Val.} &
        {\bf RBC} &
        {\bf Med.} &
        {\bf Med.} &
        {\bf P-Val.} &
        {\bf RBC} &
        {\bf Med.} &
        {\bf P-Val.} &
        {\bf RBC} \\
        \midrule 
        Unweighted & 2 & 2 & 32 & 65 & <0.01 & -1.00 & 32 & <0.01 & -0.14 & 0.83 & 0.85 & <0.01 & -0.89 & 0.79 & <0.01 & 1.00 \\
\rowcolor{gray!20}
Unweighted & 4 & 2 & 33 & 112 & <0.01 & -1.00 & 32 & <0.01 & 0.11 & 0.85 & 0.90 & <0.01 & -1.00 & 0.79 & <0.01 & 1.00 \\
Unweighted & 4 & 4 & 54 & 112 & <0.01 & -1.00 & 65 & <0.01 & -0.73 & 0.85 & 0.90 & <0.01 & -1.00 & 0.85 & <0.01 & -0.26 \\
\rowcolor{gray!20}
Unweighted & 8 & 2 & 39 & 210 & <0.01 & -1.00 & 32 & <0.01 & 0.78 & 0.91 & 0.93 & <0.01 & -0.97 & 0.79 & <0.01 & 1.00 \\
Unweighted & 8 & 4 & 57 & 210 & <0.01 & -1.00 & 65 & <0.01 & -0.57 & 0.91 & 0.93 & <0.01 & -0.96 & 0.85 & <0.01 & 0.99 \\
\rowcolor{gray!20}
Unweighted & 8 & 8 & 101 & 210 & <0.01 & -1.00 & 112 & <0.01 & -0.55 & 0.93 & 0.93 & <0.01 & -0.30 & 0.90 & <0.01 & 0.94 \\
Weighted & 2 & 2 & 31 & 65 & <0.01 & -1.00 & 32 & <0.01 & -0.32 & 0.84 & 0.85 & <0.01 & -0.77 & 0.79 & <0.01 & 1.00 \\
\rowcolor{gray!20}
Weighted & 4 & 2 & 32 & 112 & <0.01 & -1.00 & 32 & 0.07 & 0.07 & 0.87 & 0.90 & <0.01 & -0.96 & 0.79 & <0.01 & 1.00 \\
Weighted & 4 & 4 & 55 & 112 & <0.01 & -1.00 & 65 & <0.01 & -0.76 & 0.89 & 0.90 & <0.01 & -0.38 & 0.85 & <0.01 & 0.99 \\
\rowcolor{gray!20}
Weighted & 8 & 2 & 32 & 210 & <0.01 & -1.00 & 32 & <0.01 & -0.20 & 0.82 & 0.93 & <0.01 & -1.00 & 0.79 & <0.01 & 0.88 \\
Weighted & 8 & 4 & 57 & 210 & <0.01 & -1.00 & 65 & <0.01 & -0.46 & 0.89 & 0.93 & <0.01 & -1.00 & 0.85 & <0.01 & 1.00 \\
\rowcolor{gray!20}
Weighted & 8 & 8 & 103 & 210 & <0.01 & -1.00 & 112 & <0.01 & -0.49 & 0.91 & 0.93 & <0.01 & -0.95 & 0.90 & <0.01 & 0.65 \\

        \bottomrule
    \end{tabular}
\end{table*}

\begin{figure*}[t]
    \centering
    \begin{subfigure}{0.24\textwidth}
        \centering
        \includegraphics[height=25mm, width=\textwidth]{%
            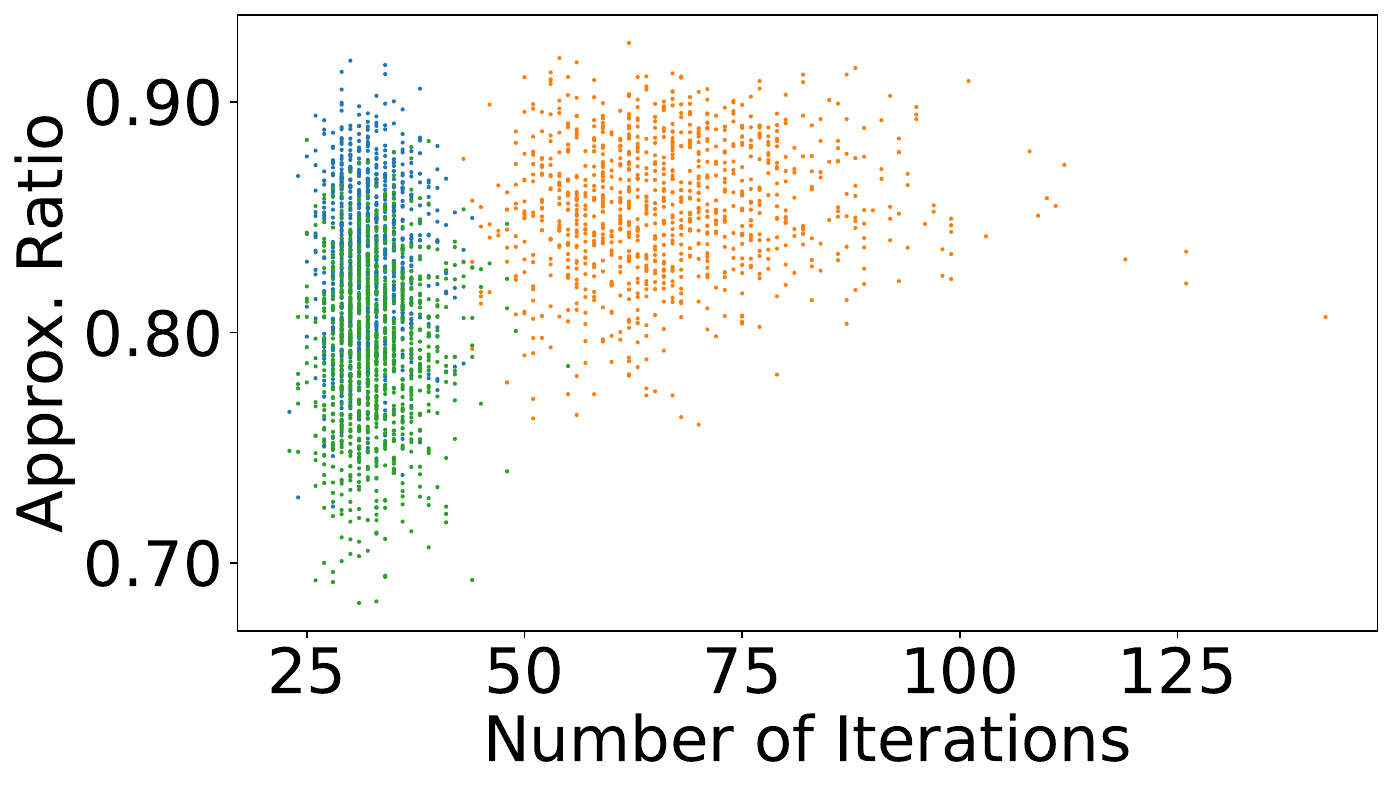
        }
        \caption{%
            \label{fig:results:unweighted22}
            Unweighted, 2 Layers, 2 Param.
        }
    \end{subfigure}
    \begin{subfigure}{0.24\textwidth}
        \centering
        \includegraphics[height=25mm, width=\textwidth]{%
            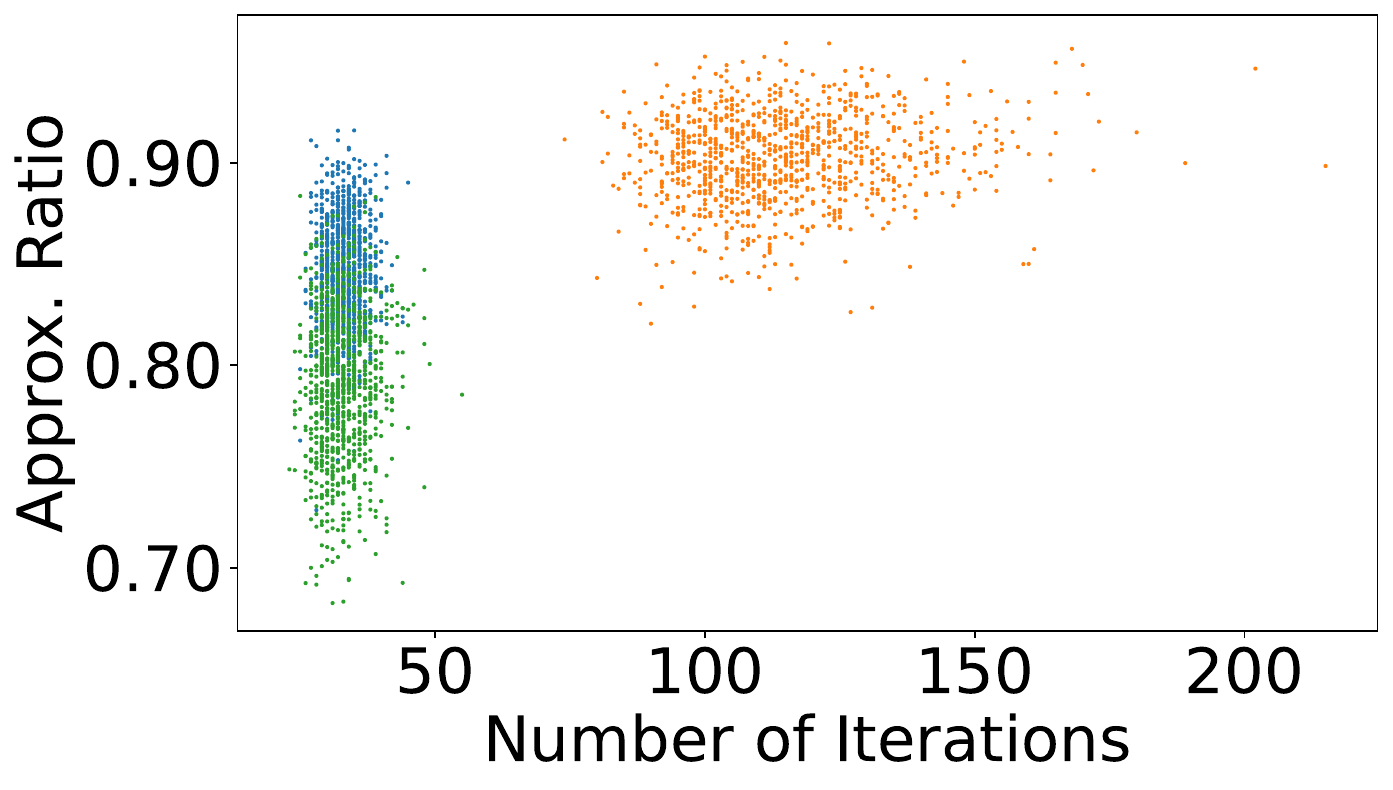
        }
        \caption{%
            \label{fig:results:unweighted42}
            Unweighted, 4 Layers, 2 Param.
        }
    \end{subfigure}
    \begin{subfigure}{0.24\textwidth}
        \centering
        \includegraphics[height=25mm, width=\textwidth]{%
            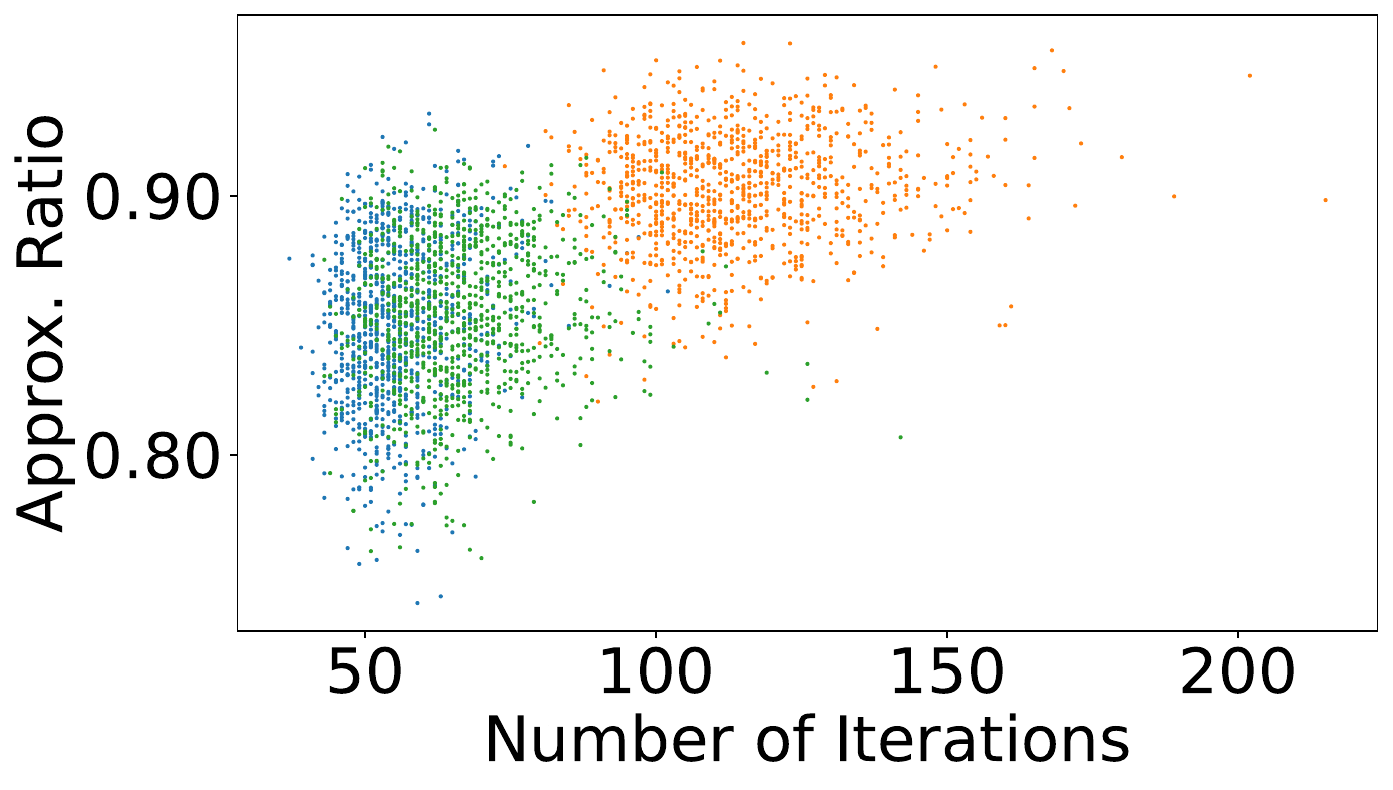
        }
        \caption{%
            \label{fig:results:unweighted44}
            Unweighted, 4 Layers, 4 Param.
        }
    \end{subfigure}
    \begin{subfigure}{0.24\textwidth}
        \centering
        \includegraphics[height=25mm, width=\textwidth]{%
            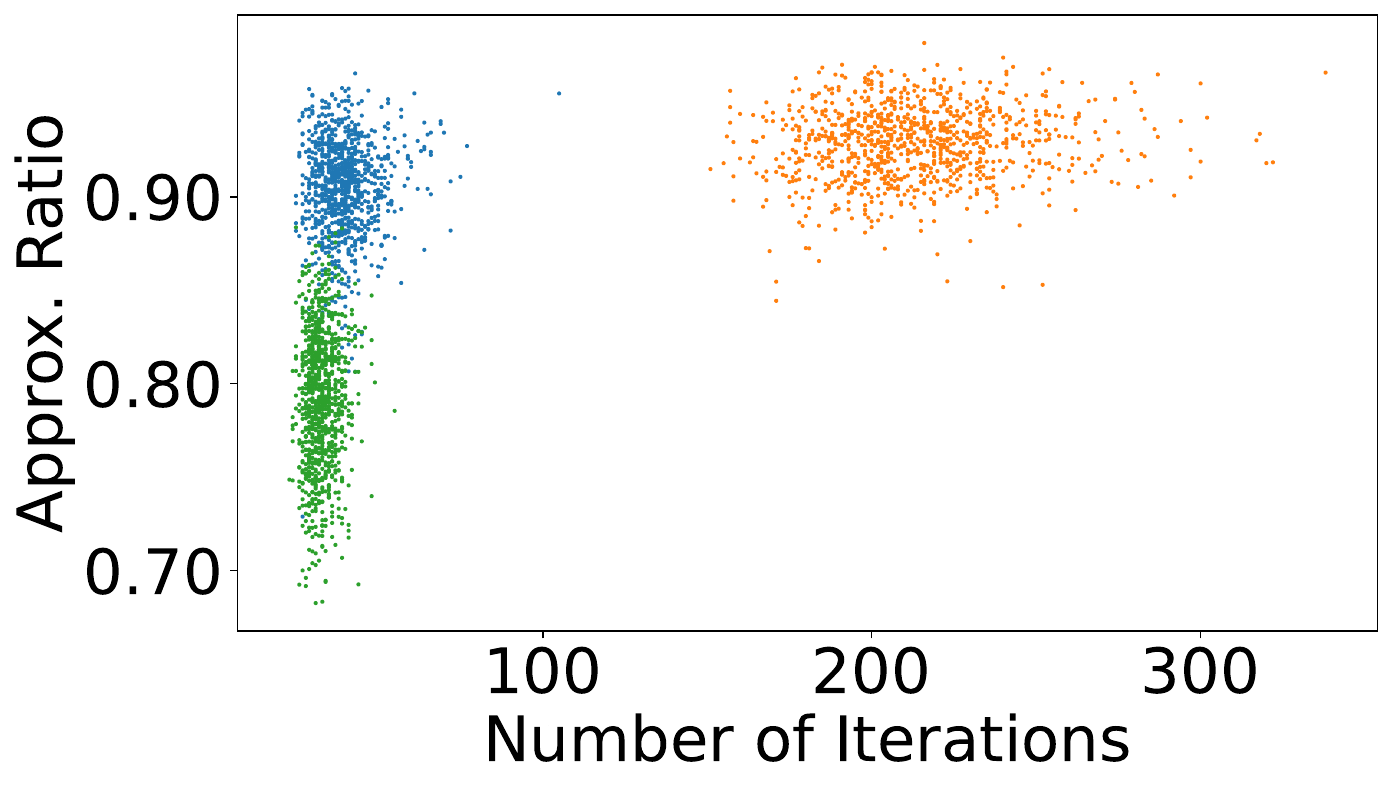
        }
        \caption{%
            \label{fig:results:unweighted82}
            Unweighted, 8 Layers, 2 Param.
        }
    \end{subfigure} \\
    \begin{subfigure}{0.24\textwidth}
        \centering
        \includegraphics[height=25mm, width=\textwidth]{%
            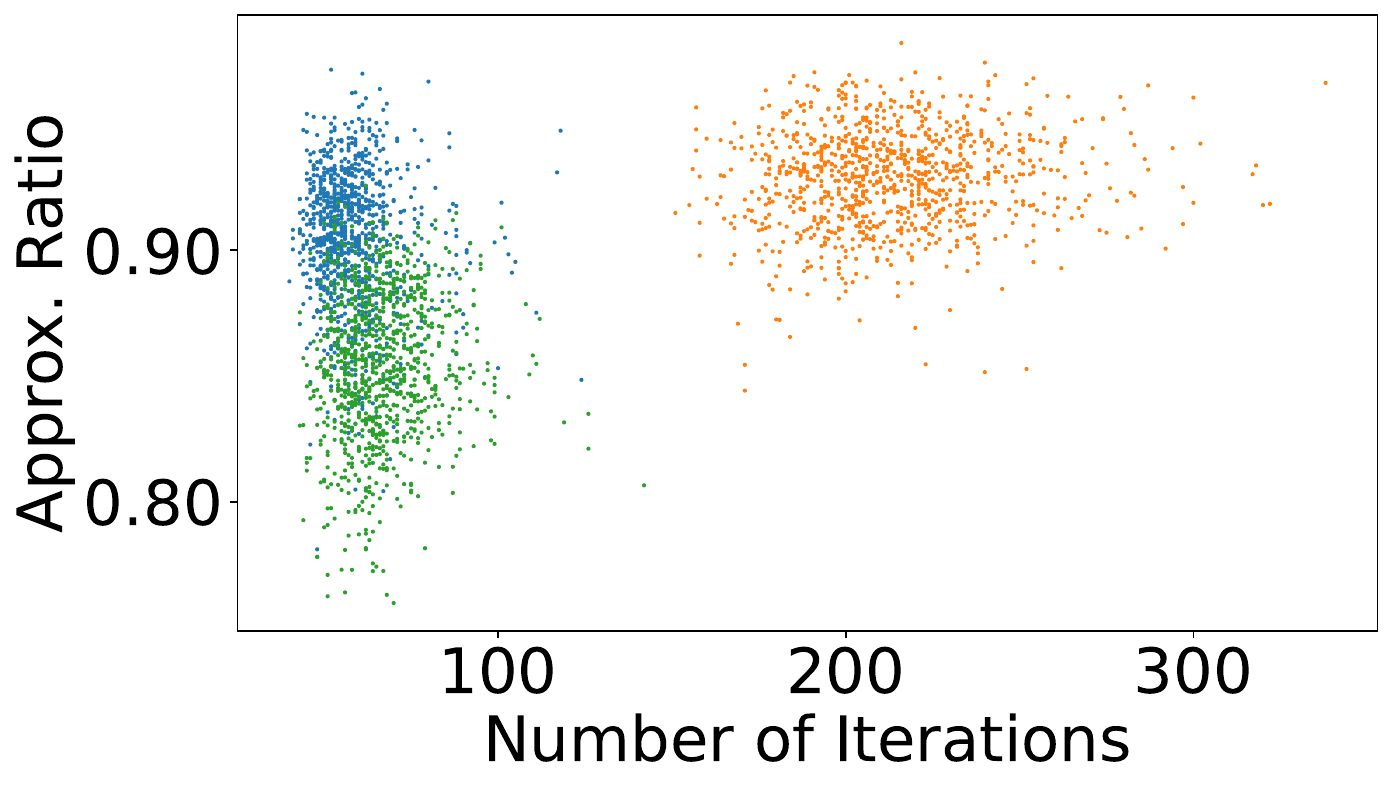
        }
        \caption{%
            \label{fig:results:unweighted84}
            Unweighted, 8 Layers, 4 Param.
        }
    \end{subfigure}
    \begin{subfigure}{0.24\textwidth}
        \centering
        \includegraphics[height=25mm, width=\textwidth]{%
            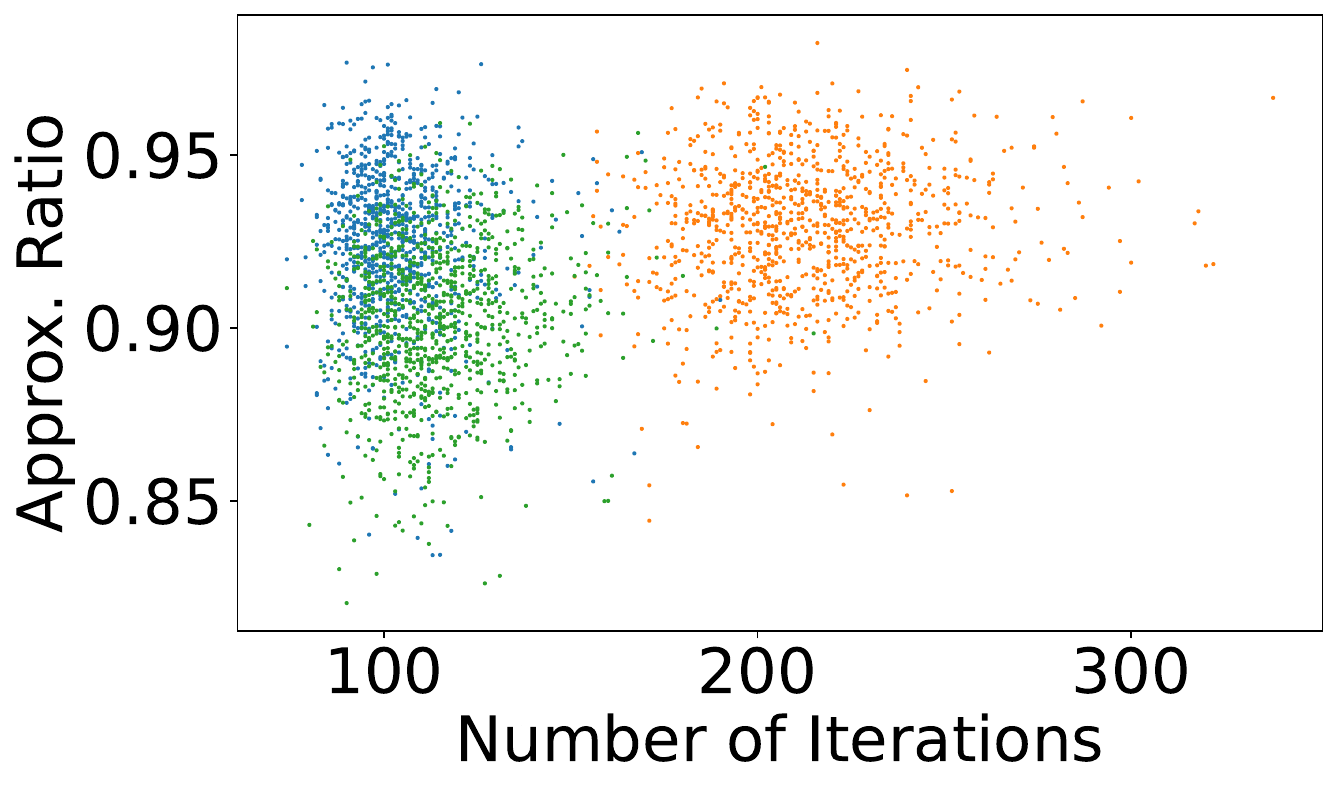
        }
        \caption{
            \label{fig:results:unweighted88}
            Unweighted, 8 Layers, 8 Param.
        }
    \end{subfigure}
    \begin{subfigure}{0.24\textwidth}
        \centering
        \includegraphics[height=25mm, width=\textwidth]{%
        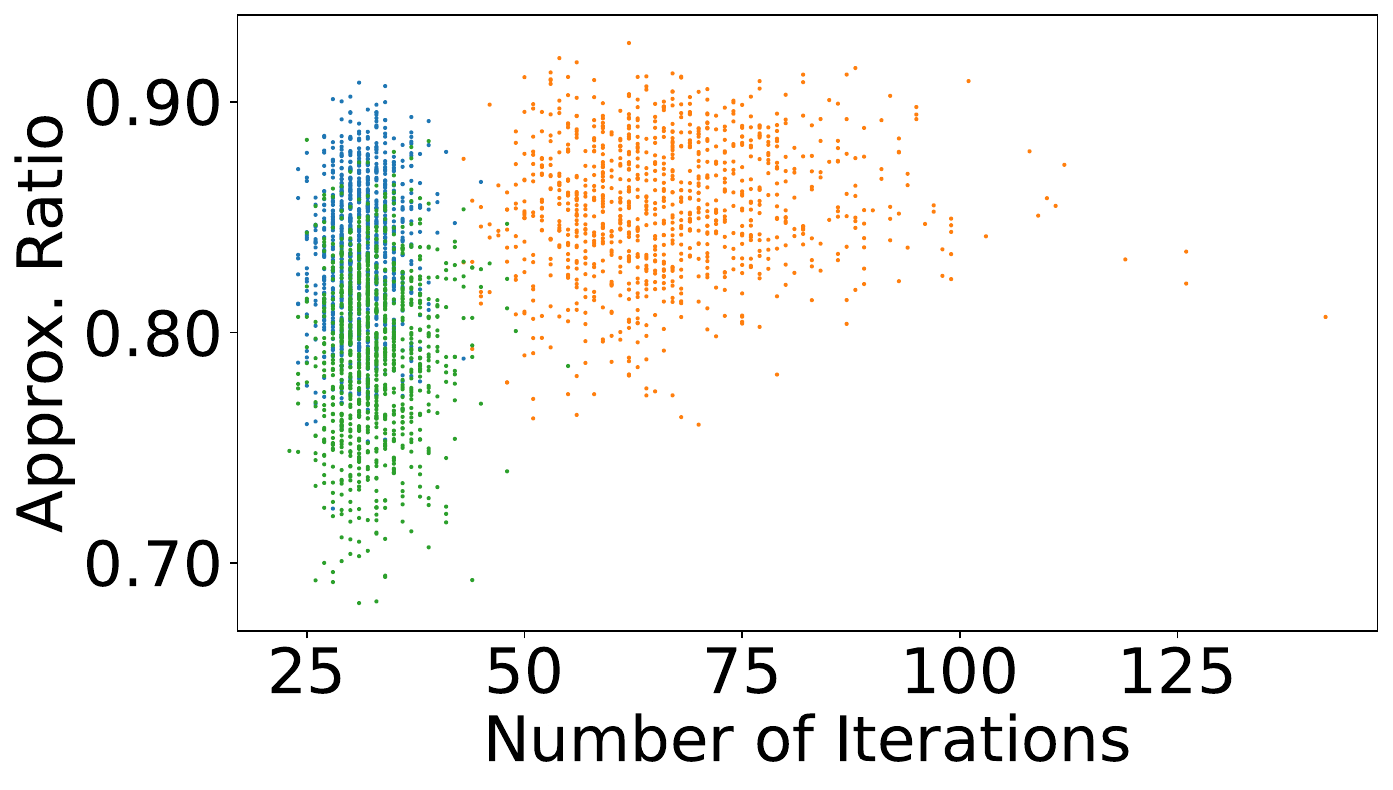
    }
        \caption{%
            \label{fig:results:weighted22}
            Weighted, 2 Layers, 2 Param.
        }
    \end{subfigure}
    \begin{subfigure}{0.24\textwidth}
        \centering
        \includegraphics[height=25mm, width=\textwidth]{%
            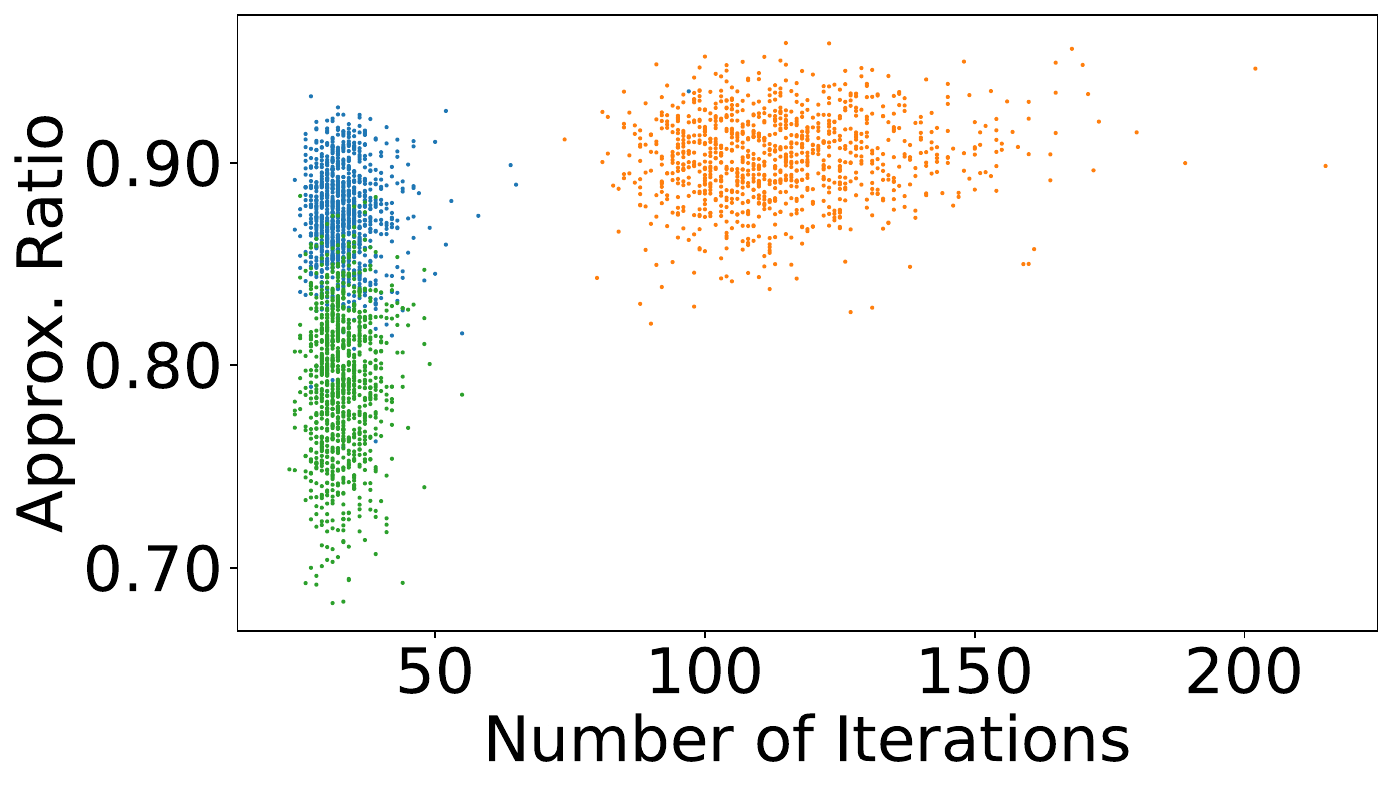
        }
        \caption{%
            \label{fig:results:weighted42}
            Weighted, 4 Layers, 2 Param.
        }
    \end{subfigure} \\
    \begin{subfigure}{0.24\textwidth}
        \centering
        \includegraphics[height=25mm, width=\textwidth]{%
            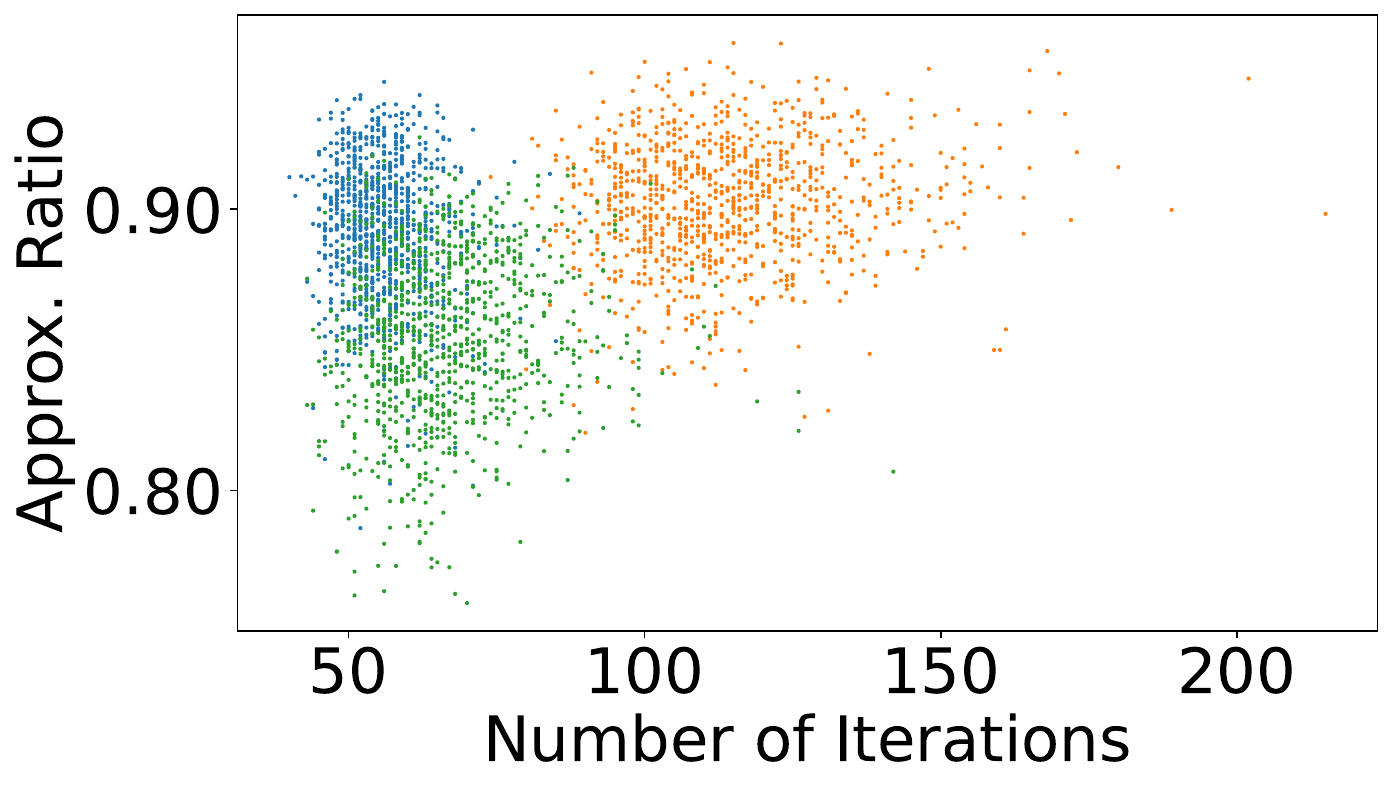
        }
        \caption{%
            \label{fig:results:weighted44}
            Weighted, 4 Layers, 4 Param.
        }
    \end{subfigure}
    \begin{subfigure}{0.24\textwidth}
        \centering
        \includegraphics[height=25mm, width=\textwidth]{%
            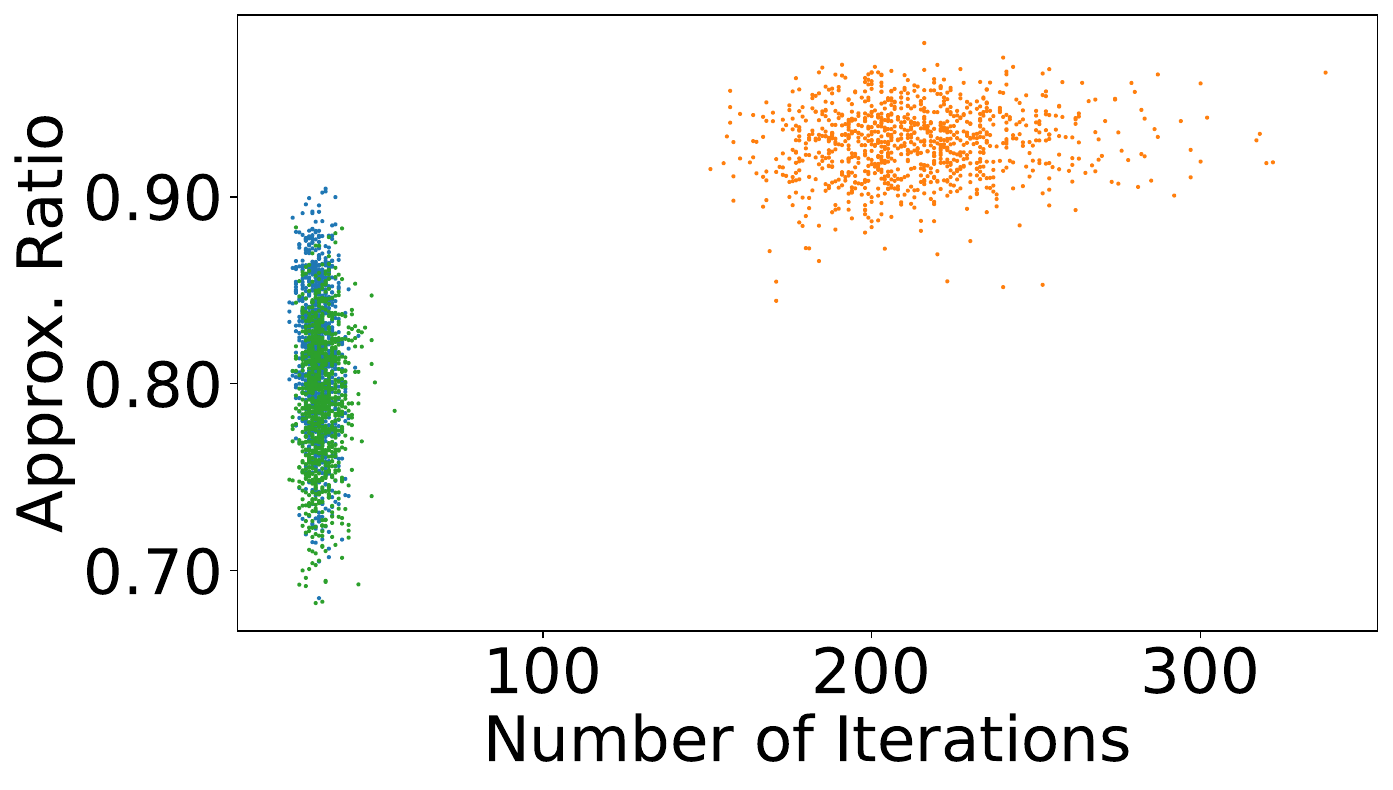
        }
        \caption{%
            \label{fig:results:weighted82}
            Weighted, 8 Layers, 2 Param.
        }
    \end{subfigure}
    \begin{subfigure}{0.24\textwidth}
        \centering
        \includegraphics[height=25mm, width=\textwidth]{%
            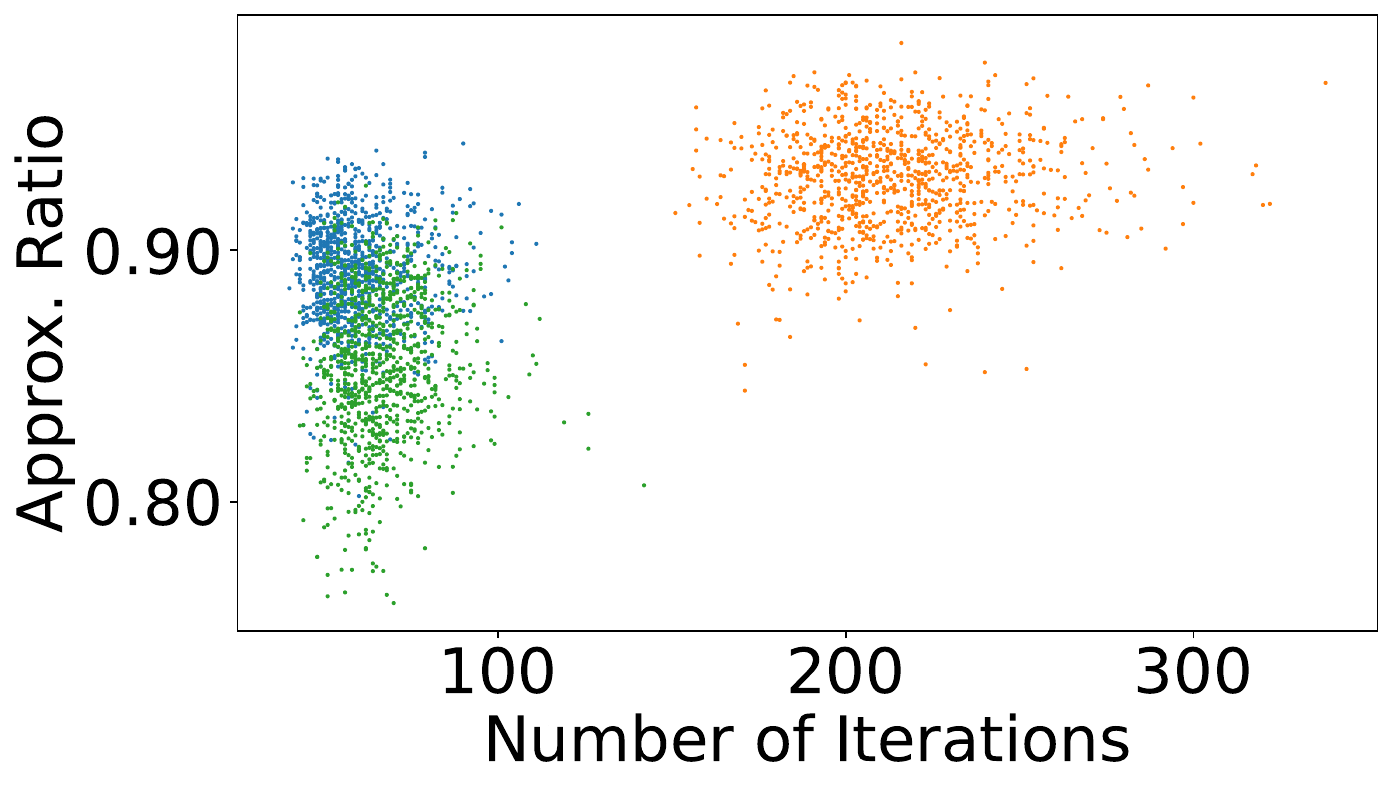
        }
        \caption{%
            \label{fig:results:weighted84}
            Weighted, 8 Layers, 4 Param.
        }
    \end{subfigure}
    \begin{subfigure}{0.24\textwidth}
        \centering
        \includegraphics[height=25mm, width=\textwidth]{%
            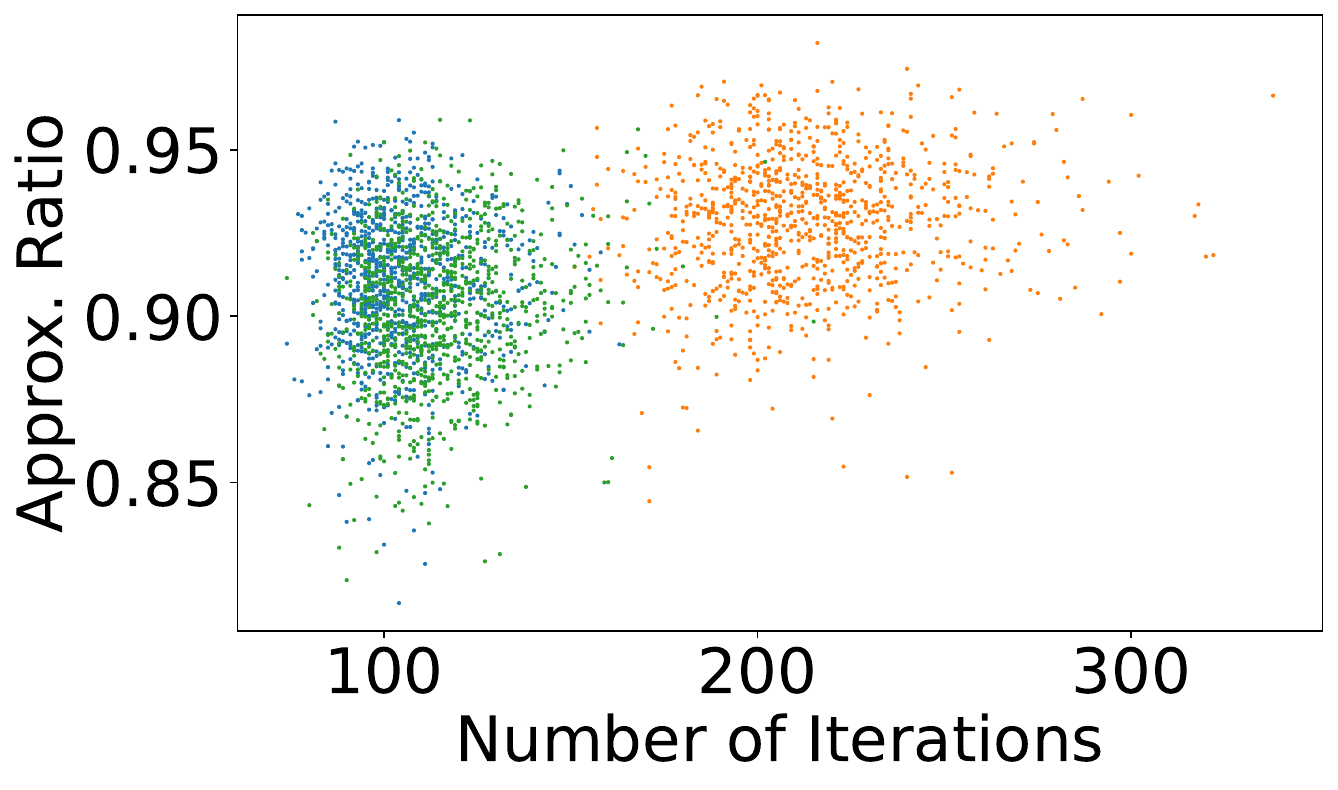
        }
        \caption{%
            \label{fig:results:weighted88}
            Weighted, 8 Layers, 8 Param.
        }
    \end{subfigure} \\
    \caption{%
        \label{fig:results}
        Distributions of the number of iterations (x-axis) and approximation 
        ratios (y-axis) for QAOA-PCA (blue), standard QAOA with the same number 
        of layers (orange), and standard QAOA with the same number of parameters 
        (green).
    }
\end{figure*}

For the 12 configurations of QAOA-PCA, Table~\ref{tab:results} gives the 
training set, the number of layers, and the number of principal components that
determines the number of parameters.
It gives the median of either the number of iterations (RQ1) or the 
approximation ratio (RQ2) attained by QAOA-PCA over the evaluation set, followed 
by the results of the comparisons against both standard QAOA baselines.
It gives the median attained by the both baselines, and the p-value and RBC 
effect sizes.
A p-value less than 0.01 indicates that we can reject the null hypothesis of
no difference at the 99\% confidence interval.
An RBC of -1 indicates QAOA-PCA always has a lower value, 1 indicates the 
opposite, and 0 indicates no difference.

Each subfigure of Figure~\ref{fig:results} corresponds to a row in 
Table~\ref{tab:results} as indicated by the captions.
They illustrate the distribution of the number of iterations (x-axis) and the
approximation ratio (y-axis) across the evaluation set for QAOA-PCA (blue), for
standard QAOA with the same number of layers (orange), and for standard QAOA 
with the same number of parameters (green).

\vspace{1mm} \noindent {\bf Results for RQ1: Efficiency.}
Table~\ref{tab:results} shows that the p-values for the comparisons between the 
number of iterations required by QAOA-PCA and standard QAOA with the same number 
of layers are significant (<0.01) for every configuration.
The effect sizes are all -1, indicating that QAOA-PCA always requires fewer
iterations.
The magnitude of this is shown by the large differences between the medians in 
favor of QAOA-PCA.
It is clear in every subfigure of Figure~\ref{fig:results} as the large gap 
along the x-axis between the data points for QAOA-PCA and standard QAOA with the 
same number of layers.

With respect to standard QAOA with the same number of parameters, all but one
of the p-values is significant.
While the majority of the effect sizes are negative, there is significant
variance in magnitude.
This is illustrated in Figure~\ref{fig:results}, where the data points for
QAOA-PCA are generally more in line along the x-axis with those of standard QAOA 
with the same number of parameters.

\conclusionbox{RQ1}{%
    QAOA-PCA consistently requires far fewer iterations than standard QAOA with
    the same layers.
    With the same number of parameters, the improvement is less pronounced.
}

\vspace{1mm} \noindent {\bf Results for RQ2: Performance.}
The difference between the approximation ratio of QAOA-PCA and standard QAOA 
with the same number of layers is always significant.
The negative effect sizes with varying magnitudes indicate QAOA-PCA often 
performs worse.
However, comparing medians, and the y-axis positions of the data points for 
QAOA-PCA and standard QAOA with the same number of layers, highlights that the 
difference is generally minor.

As for standard QAOA with the same number of parameters, the difference in
performance is once again statistically significant for every configuration.
In this case, all but one of the effect sizes are positive and are generally 
close to 1.
This indicates that QAOA-PCA very frequently performs better, as illustrated by
the generally higher y-axis positions of the data points for QAOA-PCA compared 
to those for standard QAOA with the same number of parameters.

\conclusionbox{RQ2}{%
    QAOA-PCA generally achieves slightly lower approximation ratios than
    standard QAOA with the same layers.
    However, with the same number of parameters, QAOA-PCA almost always
    outperforms standard QAOA.
}

\vspace{1mm} \noindent {\bf Discussion.}
Our results demonstrate that QAOA-PCA consistently requires fewer iterations 
than standard QAOA, achieving a substantial efficiency gain.
While this comes at the cost of a slight reduction in approximation ratio 
compared to QAOA with the same number of layers, QAOA-PCA often outperforms 
standard QAOA when matched by parameter count.
This suggests that QAOA-PCA strikes a favorable balance between efficiency and
performance, reducing optimization overhead without significantly compromising
solution quality.
This is illustrated in Figure~\ref{fig:results}, where the data points for
QAOA-PCA generally occupy the upper-left of each subfigure, which represents the 
optimum region.
Our results do not indicate any obvious differences between training on
unweighted versus weighted graphs.
This may seem surprising given the complexity of weighted MaxCut, however, 
previous research demonstrated parameter transferability between unweighted and 
weighted graphs~\cite{Shaydulin2023}.

    \section{Related Work}
\label{sec:related}

Acampora \etal proposed a clustering approach that assigns fixed parameters to 
new problem instances based on their similarity to previously addressed 
instances from a training set~\cite{Acampora2023}.
Along a similar vein, Moussa \etal investigated unsupervised machine learning
techniques, particularly clustering, to set QAOA parameters for MaxCut without
explicit optimization~\cite{Moussa2022}.
While this removes the need for classical optimization, it relies on the
assumption that new instances closely resemble those in the training set, with
no mechanism for further refinement.
Unlike these approaches, which trade all such adaptability for efficiency, 
QAOA-PCA retains some optimization capability while still reducing computational
cost.
Instead of assigning parameters from a set of clusters, QAOA-PCA applies PCA, 
another technique in unsupervised machine learning, to identify a 
lower-dimensional representation of QAOA parameters.
It reduces the number of variables that require optimization while still 
allowing adjustments for diverse unseen instances.

Amosy \etal introduced a neural network-based method for initializing QAOA 
parameters, which predicts parameters based on previously optimized instances 
with the aim of eliminating further optimization~\cite{Amosy2024}.
Their approach is effective for the problem sizes used in training, but their
study does not evaluate generalizability to larger instances.
QAOA-PCA retains the ability to refine parameters after initialization while
reducing the number of parameters, enabling it to efficiently adapt to larger 
instances.
Additionally, it provides direct control over the parameter count to adjust the 
balance between efficiency and performance.
These differences make QAOA-PCA a viable alternative when full optimization is
too costly but entirely eliminating optimization is undesirable.

    \section{Conclusion and Future Work}
\label{sec:conclusion}

Our empirical evaluation demonstrates that QAOA-PCA substantially reduces the
number of optimizer iterations.
Compared to standard QAOA, it consistently improves efficiency without
significantly compromising performance.
These results highlight the feasibility of QAOA-PCA as a promising approach in 
quantum software engineering for scaling QAOA to larger problem instances.

\vspace{1mm} \noindent {\bf Future Work.}
We aim to develop an adaptive strategy in which QAOA-PCA begins with a small 
number of principal components and expands dynamically when optimization 
progress stagnates.
This approach should reduce initial computational overhead while allowing for
increased expressivity when necessary, helping the optimizer escape plateaus and
refine solutions more effectively.
We also plan to conduct a broader empirical evaluation by applying QAOA-PCA to
larger graphs and additional problems.
We will evaluate a greater range of circuit depths, training set configurations,
and numbers of components to more rigorously determine the conditions
under which QAOA-PCA provides the best trade-off between efficiency and
performance.
Most critically, we will evaluate QAOA-PCA against noisy quantum simulators and
run experiments on real quantum hardware.
Finally, we intend to investigate alternate initialization schemes for QAOA-PCA.
In our empirical evaluation, we relied on random initialization.
We will evaluate the applicability of prior machine learning-based
approaches~\cite{Amosy2024} to produce a hybrid technique that benefits from 
both parameter reduction and data-driven starting points in the
optimization process.

    \balance
    \bibliography{bibliography}


\begin{thebibliography}{27}


\ifx \showCODEN    \undefined \def \showCODEN     #1{\unskip}     \fi
\ifx \showDOI      \undefined \def \showDOI       #1{#1}\fi
\ifx \showISBNx    \undefined \def \showISBNx     #1{\unskip}     \fi
\ifx \showISBNxiii \undefined \def \showISBNxiii  #1{\unskip}     \fi
\ifx \showISSN     \undefined \def \showISSN      #1{\unskip}     \fi
\ifx \showLCCN     \undefined \def \showLCCN      #1{\unskip}     \fi
\ifx \shownote     \undefined \def \shownote      #1{#1}          \fi
\ifx \showarticletitle \undefined \def \showarticletitle #1{#1}   \fi
\ifx \showURL      \undefined \def \showURL       {\relax}        \fi
\providecommand\bibfield[2]{#2}
\providecommand\bibinfo[2]{#2}
\providecommand\natexlab[1]{#1}
\providecommand\showeprint[2][]{arXiv:#2}

\bibitem[Rep(2025)]%
        {ReplicationPackage}
 \bibinfo{year}{2025}\natexlab{}.
\newblock \bibinfo{title}{Replication Package,
  \url{https://doi.org/10.5281/zenodo.15269564}}.
\newblock
\newblock


\bibitem[Abdi and Williams(2010)]%
        {Abdi2010}
\bibfield{author}{\bibinfo{person}{H. Abdi} {and} \bibinfo{person}{L.~J
  Williams}.} \bibinfo{year}{2010}\natexlab{}.
\newblock \showarticletitle{Principal Component Analysis}.
\newblock \bibinfo{journal}{\emph{Wiley Interdisciplinary Reviews:
  {Computational} Statistics}} (\bibinfo{year}{2010}).
\newblock


\bibitem[Acampora et~al\mbox{.}(2023)]%
        {Acampora2023}
\bibfield{author}{\bibinfo{person}{G. Acampora}, \bibinfo{person}{A. Chiatto},
  {and} \bibinfo{person}{A. Vitiello}.} \bibinfo{year}{2023}\natexlab{}.
\newblock \showarticletitle{Fuzzy Clustering for {QAOA} Complexity Reduction}.
  In \bibinfo{booktitle}{\emph{Proc. FUZZ}}. \bibinfo{pages}{1--7}.
\newblock


\bibitem[Akshay et~al\mbox{.}(2021)]%
        {Akshay2021}
\bibfield{author}{\bibinfo{person}{V. Akshay}, \bibinfo{person}{D. Rabinovich},
  \bibinfo{person}{E. Campos}, {and} \bibinfo{person}{J. Biamonte}.}
  \bibinfo{year}{2021}\natexlab{}.
\newblock \showarticletitle{Parameter Concentrations in Quantum Approximate
  Optimization}.
\newblock \bibinfo{journal}{\emph{Physical Review A}} (\bibinfo{year}{2021}).
\newblock


\bibitem[Amosy et~al\mbox{.}(2024)]%
        {Amosy2024}
\bibfield{author}{\bibinfo{person}{O. Amosy}, \bibinfo{person}{T. Danzig},
  \bibinfo{person}{O. Lev}, \bibinfo{person}{E. Porat}, \bibinfo{person}{G.
  Chechik}, {and} \bibinfo{person}{A. Makmal}.}
  \bibinfo{year}{2024}\natexlab{}.
\newblock \showarticletitle{Iteration-Free Quantum Approximate Optimization
  Algorithm Using Neural Networks}.
\newblock \bibinfo{journal}{\emph{Quantum Machine Intelligence}}
  (\bibinfo{year}{2024}).
\newblock


\bibitem[Brandao et~al\mbox{.}(2018)]%
        {Brandao2018}
\bibfield{author}{\bibinfo{person}{F.~G. S.~L. Brandao}, \bibinfo{person}{M.
  Broughton}, \bibinfo{person}{E. Farhi}, \bibinfo{person}{S. Gutmann}, {and}
  \bibinfo{person}{H. Neven}.} \bibinfo{year}{2018}\natexlab{}.
\newblock \showarticletitle{For Fixed Control Parameters the Quantum
  Approximate Optimization Algorithm's Objective Function Value Concentrates
  for Typical Instances}.
\newblock \bibinfo{journal}{\emph{arXiv preprint arXiv:1812.04170}}
  (\bibinfo{year}{2018}).
\newblock


\bibitem[Campbell and Dahl(2022)]%
        {Campbell2022}
\bibfield{author}{\bibinfo{person}{C. Campbell} {and} \bibinfo{person}{E.
  Dahl}.} \bibinfo{year}{2022}\natexlab{}.
\newblock \showarticletitle{{QAOA} of the Highest Order}. In
  \bibinfo{booktitle}{\emph{Proc. ICSA-C}}. \bibinfo{pages}{141--146}.
\newblock


\bibitem[Farhi et~al\mbox{.}(2014)]%
        {Farhi2014}
\bibfield{author}{\bibinfo{person}{E. Farhi}, \bibinfo{person}{J. Goldstone},
  {and} \bibinfo{person}{S. Gutmann}.} \bibinfo{year}{2014}\natexlab{}.
\newblock \showarticletitle{A Quantum Approximate Optimization Algorithm}.
\newblock \bibinfo{journal}{\emph{arXiv preprint arXiv:1411.4028}}
  (\bibinfo{year}{2014}).
\newblock


\bibitem[Galda et~al\mbox{.}(2023)]%
        {Galda2023}
\bibfield{author}{\bibinfo{person}{A. Galda}, \bibinfo{person}{E. Gupta},
  \bibinfo{person}{J. Falla}, \bibinfo{person}{X. Liu}, \bibinfo{person}{D.
  Lykov}, \bibinfo{person}{Y. Alexeev}, {and} \bibinfo{person}{I. Safro}.}
  \bibinfo{year}{2023}\natexlab{}.
\newblock \showarticletitle{Similarity-Based Parameter Transferability in the
  Quantum Approximate Optimization Algorithm}.
\newblock \bibinfo{journal}{\emph{Frontiers in Quantum Science and Technology}}
  (\bibinfo{year}{2023}).
\newblock


\bibitem[Galda et~al\mbox{.}(2021)]%
        {Galda2021}
\bibfield{author}{\bibinfo{person}{A. Galda}, \bibinfo{person}{X. Liu},
  \bibinfo{person}{D. Lykov}, \bibinfo{person}{Y. Alexeev}, {and}
  \bibinfo{person}{I. Safro}.} \bibinfo{year}{2021}\natexlab{}.
\newblock \showarticletitle{Transferability of Optimal {QAOA} Parameters
  Between Random Graphs}. In \bibinfo{booktitle}{\emph{Proc. QCE}}.
  \bibinfo{pages}{171--180}.
\newblock


\bibitem[Hao et~al\mbox{.}(2024)]%
        {Hao2024}
\bibfield{author}{\bibinfo{person}{T. Hao}, \bibinfo{person}{Z. He},
  \bibinfo{person}{R. Shaydulin}, \bibinfo{person}{J. Larson}, {and}
  \bibinfo{person}{M. Pistoia}.} \bibinfo{year}{2024}\natexlab{}.
\newblock \showarticletitle{End-to-End Protocol for High-Quality {QAOA}
  Parameters With Few Shots}.
\newblock \bibinfo{journal}{\emph{arXiv preprint arXiv:2408.00557}}
  (\bibinfo{year}{2024}).
\newblock


\bibitem[He et~al\mbox{.}(2024)]%
        {He2024}
\bibfield{author}{\bibinfo{person}{Z. He}, \bibinfo{person}{R. Shaydulin},
  \bibinfo{person}{D. Herman}, \bibinfo{person}{C. Li}, \bibinfo{person}{R.
  Raymond}, \bibinfo{person}{S.~H. Sureshbabu}, {and} \bibinfo{person}{M.
  Pistoia}.} \bibinfo{year}{2024}\natexlab{}.
\newblock \showarticletitle{Parameter Setting Heuristics Make the Quantum
  Approximate Optimization Algorithm Suitable for the Early Fault-Tolerant
  Era}.
\newblock \bibinfo{journal}{\emph{arXiv preprint arXiv:2408.09538}}
  (\bibinfo{year}{2024}).
\newblock


\bibitem[Javadi-Abhari et~al\mbox{.}(2024)]%
        {Javadi2024}
\bibfield{author}{\bibinfo{person}{A. Javadi-Abhari}, \bibinfo{person}{M.
  Treinish}, \bibinfo{person}{K. Krsulich}, \bibinfo{person}{C.~J. Wood},
  \bibinfo{person}{J. Lishman}, \bibinfo{person}{J. Gacon}, \bibinfo{person}{S.
  Martiel}, \bibinfo{person}{P.~D. Nation}, \bibinfo{person}{L.~S. Bishop},
  \bibinfo{person}{A.~W. Cross}, \bibinfo{person}{Johnson~B. R.}, {and}
  \bibinfo{person}{Gambetta~J. M.}} \bibinfo{year}{2024}\natexlab{}.
\newblock \showarticletitle{Quantum Computing With {Qiskit}}.
\newblock \bibinfo{journal}{\emph{arXiv preprint arXiv:2405.08810}}
  (\bibinfo{year}{2024}).
\newblock


\bibitem[Lyngfelt and Garc{\'\i}a-{\'A}lvarez(2025)]%
        {Lyngfelt2025}
\bibfield{author}{\bibinfo{person}{I. Lyngfelt} {and} \bibinfo{person}{L.
  Garc{\'\i}a-{\'A}lvarez}.} \bibinfo{year}{2025}\natexlab{}.
\newblock \showarticletitle{Symmetry-Informed Transferability of Optimal
  Parameters in the Quantum Approximate Optimization Algorithm}.
\newblock \bibinfo{journal}{\emph{Physical Review A}} (\bibinfo{year}{2025}).
\newblock


\bibitem[McKay(1983)]%
        {McKay1983}
\bibfield{author}{\bibinfo{person}{B.~D. McKay}.}
  \bibinfo{year}{1983}\natexlab{}.
\newblock \showarticletitle{Applications of a Technique for Labelled
  Enumeration}.
\newblock \bibinfo{journal}{\emph{Congressus Numerantium}}
  (\bibinfo{year}{1983}).
\newblock


\bibitem[Montanez-Barrera et~al\mbox{.}(2024)]%
        {Montanez2024}
\bibfield{author}{\bibinfo{person}{J.~A. Montanez-Barrera}, \bibinfo{person}{D.
  Willsch}, {and} \bibinfo{person}{K. Michielsen}.}
  \bibinfo{year}{2024}\natexlab{}.
\newblock \showarticletitle{Transfer Learning of Optimal {QAOA} Parameters in
  Combinatorial Optimization}.
\newblock \bibinfo{journal}{\emph{arXiv preprint arXiv:2402.05549}}
  (\bibinfo{year}{2024}).
\newblock


\bibitem[Moussa et~al\mbox{.}(2022)]%
        {Moussa2022}
\bibfield{author}{\bibinfo{person}{C. Moussa}, \bibinfo{person}{H. Wang},
  \bibinfo{person}{T. B{\"a}ck}, {and} \bibinfo{person}{V. Dunjko}.}
  \bibinfo{year}{2022}\natexlab{}.
\newblock \showarticletitle{Unsupervised Strategies for Identifying Optimal
  Parameters in Quantum Approximate Optimization Algorithm}.
\newblock \bibinfo{journal}{\emph{EPJ Quantum Technology}}
  (\bibinfo{year}{2022}).
\newblock


\bibitem[Nielsen and Chuang(2010)]%
        {Nielsen2010}
\bibfield{author}{\bibinfo{person}{M.~A. Nielsen} {and} \bibinfo{person}{I.~L.
  Chuang}.} \bibinfo{year}{2010}\natexlab{}.
\newblock \bibinfo{booktitle}{\emph{Quantum Computation and Quantum
  Information}}.
\newblock


\bibitem[Pellow-Jarman et~al\mbox{.}(2021)]%
        {Pellow2021}
\bibfield{author}{\bibinfo{person}{A. Pellow-Jarman}, \bibinfo{person}{I.
  Sinayskiy}, \bibinfo{person}{A. Pillay}, {and} \bibinfo{person}{F.
  Petruccione}.} \bibinfo{year}{2021}\natexlab{}.
\newblock \showarticletitle{A Comparison of Various Classical Optimizers for a
  Variational Quantum Linear Solver}.
\newblock \bibinfo{journal}{\emph{Quantum Information Processing}}
  (\bibinfo{year}{2021}).
\newblock


\bibitem[Preskill(2018)]%
        {Preskill2018}
\bibfield{author}{\bibinfo{person}{J. Preskill}.}
  \bibinfo{year}{2018}\natexlab{}.
\newblock \showarticletitle{Quantum Computing in the {NISQ} Era and Beyond}.
\newblock \bibinfo{journal}{\emph{Quantum}} (\bibinfo{year}{2018}).
\newblock


\bibitem[Sack and Serbyn(2021)]%
        {Sack2021}
\bibfield{author}{\bibinfo{person}{S.~H. Sack} {and} \bibinfo{person}{M.
  Serbyn}.} \bibinfo{year}{2021}\natexlab{}.
\newblock \showarticletitle{Quantum Annealing Initialization of the Quantum
  Approximate Optimization Algorithm}.
\newblock \bibinfo{journal}{\emph{Quantum}} (\bibinfo{year}{2021}).
\newblock


\bibitem[Schw{\"a}gerl et~al\mbox{.}(2024)]%
        {Schwagerl2024}
\bibfield{author}{\bibinfo{person}{T. Schw{\"a}gerl}, \bibinfo{person}{Y.
  Chai}, \bibinfo{person}{T. Hartung}, \bibinfo{person}{K. Jansen}, {and}
  \bibinfo{person}{S. K{\"u}hn}.} \bibinfo{year}{2024}\natexlab{}.
\newblock \showarticletitle{Benchmarking Variational Quantum Algorithms for
  Combinatorial Optimization in Practice}.
\newblock \bibinfo{journal}{\emph{arXiv preprint arXiv:2408.03073}}
  (\bibinfo{year}{2024}).
\newblock


\bibitem[Shaydulin et~al\mbox{.}(2023)]%
        {Shaydulin2023}
\bibfield{author}{\bibinfo{person}{R. Shaydulin}, \bibinfo{person}{P.~C.
  Lotshaw}, \bibinfo{person}{J. Larson}, \bibinfo{person}{J. Ostrowski}, {and}
  \bibinfo{person}{T.~S. Humble}.} \bibinfo{year}{2023}\natexlab{}.
\newblock \showarticletitle{Parameter Transfer for Quantum Approximate
  Optimization of Weighted {MaxCut}}.
\newblock \bibinfo{journal}{\emph{Transactions on Quantum Computing}}
  (\bibinfo{year}{2023}).
\newblock


\bibitem[Shi et~al\mbox{.}(2022)]%
        {Shi2022}
\bibfield{author}{\bibinfo{person}{K. Shi}, \bibinfo{person}{R. Herrman},
  \bibinfo{person}{R. Shaydulin}, \bibinfo{person}{S. Chakrabarti},
  \bibinfo{person}{M. Pistoia}, {and} \bibinfo{person}{J. Larson}.}
  \bibinfo{year}{2022}\natexlab{}.
\newblock \showarticletitle{Multiangle {QAOA} Does Not Always Need All Its
  Angles}. In \bibinfo{booktitle}{\emph{Proc. SEC}}. \bibinfo{pages}{414--419}.
\newblock


\bibitem[Sureshbabu et~al\mbox{.}(2024)]%
        {Sureshbabu2024}
\bibfield{author}{\bibinfo{person}{S.~H. Sureshbabu}, \bibinfo{person}{D.
  Herman}, \bibinfo{person}{R. Shaydulin}, \bibinfo{person}{J. Basso},
  \bibinfo{person}{S. Chakrabarti}, \bibinfo{person}{Y. Sun}, {and}
  \bibinfo{person}{M. Pistoia}.} \bibinfo{year}{2024}\natexlab{}.
\newblock \showarticletitle{Parameter Setting in Quantum Approximate
  Optimization of Weighted Problems}.
\newblock \bibinfo{journal}{\emph{Quantum}} (\bibinfo{year}{2024}).
\newblock


\bibitem[W. and W.(2010)]%
        {Wang2010}
\bibfield{author}{\bibinfo{person}{Rui-Sheng W.} {and} \bibinfo{person}{Li-Min
  W.}} \bibinfo{year}{2010}\natexlab{}.
\newblock \showarticletitle{Maximum Cut in Fuzzy Nature: {Models} and
  Algorithms}.
\newblock \bibinfo{journal}{\emph{J. Comput. Appl. Math.}}
  (\bibinfo{year}{2010}).
\newblock


\bibitem[Zeng et~al\mbox{.}(2024)]%
        {Zeng2024}
\bibfield{author}{\bibinfo{person}{H. Zeng}, \bibinfo{person}{F. Meng},
  \bibinfo{person}{T. Luan}, \bibinfo{person}{X. Yu}, {and} \bibinfo{person}{Z.
  Zhang}.} \bibinfo{year}{2024}\natexlab{}.
\newblock \showarticletitle{Improved Quantum Approximate Optimization Algorithm
  for Low-Density Parity-Check Channel Decoding}.
\newblock \bibinfo{journal}{\emph{Advanced Quantum Technologies}}
  (\bibinfo{year}{2024}).
\newblock


\end{thebibliography}
\end{document}